\documentclass[runningheads]{llncs}
\usepackage{graphicx}

\usepackage{tikz}
\usepackage{comment}
\usepackage{amsmath,amssymb} 
\usepackage{color}
\usepackage{booktabs}
\usepackage{makecell}
\usepackage{hyperref}
\usepackage[width=122mm,left=12mm,paperwidth=146mm,height=193mm,top=12mm,paperheight=217mm]{geometry}

\begin{document}
\pagestyle{headings}
\mainmatter
\def\ECCVSubNumber{2852}  

\title{Learning from Multiple Annotator Noisy Labels via Sample-wise Label Fusion} 

\titlerunning{Learning from Multiple Annoator Noisy Labels}
%
\author{Zhengqi Gao\inst{1} \and
Fan-Keng Sun\inst{1} \and
Mingran Yang\inst{1} \and\\
Sucheng Ren\inst{2} \and
Zikai Xiong\inst{1} \and
Marc Engeler\inst{3} \and
Antonio Burazer\inst{3} \and
Linda Wildling\inst{3} \and
Luca Daniel\inst{1} \and
Duane S. Boning\inst{1}}
\authorrunning{Z. Gao et al.}
%
\institute{Massachusetts Institute of Technology, Cambridge MA 02139, USA\\ 
\and
South China University of Technology, Guangzhou, China\\
\and
Takeda Pharmaceuticals Co., Ltd., Zurich, Switzerland\\
}
\maketitle

\begin{abstract}
 Data lies at the core of modern deep learning. The impressive performance of supervised learning is built upon a base of massive accurately labeled data. However, in some real-world applications, accurate labeling might not be viable; instead, multiple noisy labels (instead of one accurate label) are provided by several annotators for each data sample. Learning a classifier on such a noisy training dataset is a challenging task. Previous approaches usually assume that all data samples share the same set of parameters related to annotator errors, while we demonstrate that label error learning should be both annotator and data sample dependent. Motivated by this observation, we propose a novel learning algorithm. The proposed method displays superiority compared with several state-of-the-art baseline methods on MNIST, CIFAR-100, and ImageNet-100. Our code is available at: \href{https://github.com/zhengqigao/Learning-from-Multiple-Annotator-Noisy-Labels}{https://github.com/zhengqigao/Learning-from-Multiple-Annotator-Noisy-Labels}.
\end{abstract}

\section{Introduction}\label{sec:intro}

In addition to improved neural network architectures (e.g., residual connections~\cite{resnet}, batch-norm~\cite{batchnorm}), the prevalence and success of modern deep learning are attributed to the availability of large datasets (e.g., CIFAR-100~\cite{cifar}, COCO~\cite{coco}, ImagNet~\cite{lifeifei-imagenet}). The massive amount of labeled data plays a critical role in the training of deep neural networks under a supervised learning setting. Thanks to the efforts of many researchers, access to these accurately labeled data is so convenient that we often take them for granted.


However, in many real-world applications, large numbers of accurate labels are not available or practicable to generate. Instead, only multiple noisy labels for data samples are gathered, due to economic limitations. Consider the need for training data to support a binary labeling (i.e., good or bad) neural network of drug vial images in an automated visual inspection system for pharmaceutical products. Training dataset generation involves labeling of a set of collected vial images (with and without defects) by a modest number of highly trained human experts. Because visual acuity, conditions, and expertise can vary among experts, not all experts will agree on labels. As a result, golden labels are not always available. Instead, we seek to take maximal advantage of the available labels from these multiple annotators.
 
This problem corresponds to a supervised classification task on a training dataset with multiple noisy labels available for each data point. To address this problem, a naive approach is to aggregate labels for each sample via a weighted summation, followed by a vanilla training procedure on the weighted aggregate. When all elements of the weight vector are equal, this approach is known as majority voting~\cite{bayes2_mjv}. More sophisticated approaches have been proposed over the past several decades. The first major category, built on Bayesian methods~\cite{bayes1,bayes2_mjv,bayes3}, has been dominant before the era of deep learning. In these methods, a probabilistic model is first defined and the maximum likelihood estimation or the maximum-a-posteriori solution is found by leveraging the expectation maximization (EM) algorithm. The second category, which is learning-based, has emerged more recently. For instance, WDN~\cite{wdn-aaai} generalizes the idea of majority voting by learning the \textit{weight vector} instead of directly setting all its entries to a single constant. In contrast, rather than using a weight vector, MBEM~\cite{mbem} introduces an annotator-specific \textit{confusion matrix} to mimic the labeling process of each annotator, and embeds this into the training of the classifier using an EM framework. Later, the authors in~\cite{cvpr} proposed to learn the annotator confusion matrices by a novel loss function involving trace regularization (referred to as TraceReg in our paper).

However, these previous methods adopt assumptions -- that all data samples share a single annotator weight vector, or that the same set of confusion matrices apply to all data samples -- that can be overly limiting, as will be demonstrated in this paper. Motivated by this observation, we propose a new learning algorithm to jointly learn sample-wise weight vectors and sample-wise annotator confusion matrices, and thus make label fusion possible. Specifically, for any input data sample, our neural network outputs its label prediction, an annotator weight vector, and a set of confusion matrices. To carry out the training process to learn these values, a novel loss function is proposed. Furthermore, for practical utility, we take advantage of the Birkhoff–von Neumann theorem and matrix decomposition technique so that only a small set of coefficients can be learned to approximate the set of confusion matrices. To exhibit our method's superior performance, we compare it with several state-of-the-art baseline methods on MNIST, CIFAR-100, and ImageNet-100.

\section{Preliminaries}
\subsection{Related Works}

\textbf{Multi-Label Classification.} Generally, multi-label classification refers to the classification problem where multiple (valid) labels are assigned to each instance. For example, in the movie genre classification problem, one movie could belong to each of action, comedy, and fiction classes at the same time. One intuitive way to solve this problem is converting to several separate binary classification problems. Specifically, if there are $R$ distinct classes that one instance could belong to, we assign a set of binary labels $\{y^{(r)}\in\{0,1\}\}_{r=1}^R$ to the instance $\mathbf{x}$, where $R$ is the total number of classes and the $r$-th label represents whether the sample belongs to the $r$-th class or not. The same formalism can be adopted in our problem, but we now interpret $R$ as the number of annotators and notice that $y^{(r)}$ resides in $\{0,1,\cdots,K-1\}$ (instead of $\{0,1\}$) for a $K$-class classification problem. One subtle difference is that in our problem each instance has only one correct label, i.e., ideally, the provided labels $\{y^{(r)}\}_{r=1}^R$ should be identical for a specific sample $\mathbf{x}$ (i.e., $y^{(1)}=y^{(2)}=\cdots=y^{(R)}$). However, in multi-label classification, $y^{(1)}$ could be different from $y^{(2)}$ in principle, with both labels being correct.

\paragraph{\textbf{\textup{Learning with Noisy Labels.}}} In a conventional $K$-class classification setting, `noisy label' refers to the fact that the label $y$ assigned to the instance $\mathbf{x}$ might be corrupted. Learning with noisy labels has been a hot topic for the past several years and various methods have been proposed~\cite{noisy_label,noisy_label2,noisy_label3,noisy_label4,noisy_label5,noisy_label6,noisy_label7,noisy_label8,noisy_label9}. These methods can be generally grouped into two categories~\cite{noisy_label}: (i) model-based approaches, and (ii) model-free approaches. Model-based approaches attempt to find the underlying noise structure and eliminate its impact from the observed data, while model-free approaches aim to achieve label noise robustness without explicitly modeling the noise~\cite{noisy_label}. To name one example for each kind, the noisy channel approach~\cite{noisy_label3} assumes a noisy channel on top of a base classifier~\cite{noisy_label}. It will learn the noise structure in the training phase and thus the base classifier can be trained using the processed clean labels. On the other hand, many model-free approaches focus on designing loss functions robust to the noise~\cite{noisy_label8,noisy_label9}. For instance, the authors in~\cite{noisy_label9} show that mean absolute value of error (MAE) is more resilient to noise, compared with the commonly used categorical cross entropy loss. 

In our problem, we have multiple noisy labels provided by different annotators which might be consistent or inconsistent. Majority voting~\cite{bayes2_mjv} simply aggregates all labels via a weighted summation over one-hot-encoded annotator labels with a $1/R$ constant weight vector, while WDN~\cite{wdn-aaai} automatically learns the weight vector. On the other hand, MBEM~\cite{mbem} and TraceReg~\cite{cvpr} both introduce the concept of a per-annotator confusion matrix to model the labeling of each annotator. In what follows, we demonstrate that the data sample-independent assumption used in these methods is overly restrictive, and justify why an alternative data sample-dependent configuration is intriguing.

\subsection{Motivations}
 Formally, we consider a supervised $K$-class classification problem on a given dataset $\mathcal{D}=\{(\mathbf{x}_n,{y}_n^{(r)})\,|\,n=1,2,\cdots, N, r = 1,2,\cdots, R\}$, where $(\mathbf{x}_n,y_n^{(r)} )$ denotes the $n$-th input feature and its corresponding label from the $r$-th annotator, $N$ and $R$ respectively represent the number of samples and annotators. In our paper, when the bold symbol $\mathbf{y}_n^{(r)}\in\mathbb{R}^K$ is presented, it denotes the one-hot-encoding of $y_n^{(r)}\in\{0,1,\cdots,K-1\}$.
 
\begin{figure}[!ht ]
    \centering
    \includegraphics[width=1.0\textwidth]{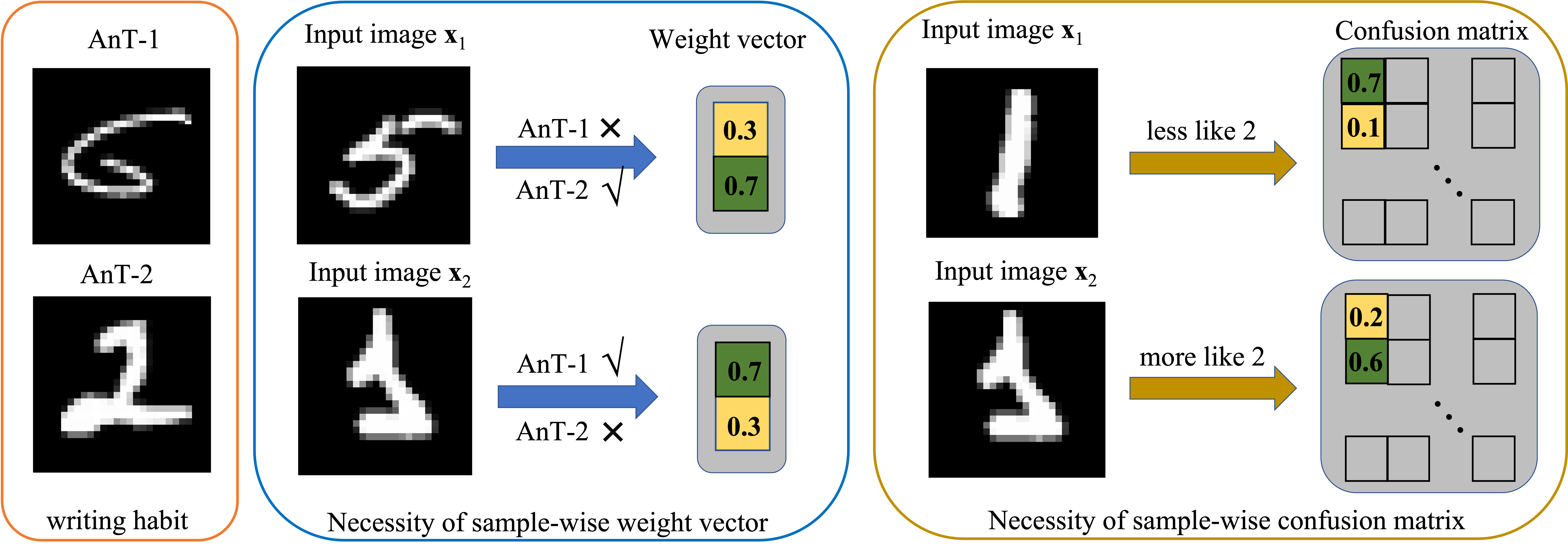}
    \caption{An illustration of why weight vector and confusion matrix need to be sample-wise using MNIST dataset. (i) The orange box contains examples of the first and second annotator writing `6' and `2', respectively. (ii) In the blue box, when input image $\mathbf{x}_1$ is provided, AnT-1 might assign label `6' due to his/her writing habit, while AnT-2 provides the correct label $5$. Thus in this case, the weight vector should emphasize on AnT-2. Alternatively, the weight vector should emphasize on AnT-1 when $\mathbf{x}_2$ is given. (iii) In the yellow box, we only consider AnT-2. When $\mathbf{x}_1$ is given, it looks less like digit `2', thus ${P}_{21}^{(2)}=\text{Pr}[y_1^{(2)}=2|y_1^{\star}=1]$ is small (say $0.1$). Yet when $\mathbf{x}_2$ is provided, ${P}_{21}^{(2)}=\text{Pr}[y_1^{(2)}=2|y_1^{\star}=1]$ will be large (say $0.6$).}
    \label{fig:motivation}
\end{figure}

 Two sorts of parameters are usually exploited in previous relevant work. The first one is a weight vector $\mathbf{w}\in\mathbb{R}^R$ used by majority voting and WDN~\cite{wdn-aaai}. Namely, they use $\mathbf{w}$ to weight the opinions of annotators and assume that the golden (one-hot-encoded) label $\mathbf{y}_n^{\star}$ can be approximated by a weighted summation of all annotators' labels $\mathbf{y}_n^{\text{targ}}$:
\begin{equation}
    \mathbf{y}_n^{\star}\approx \mathbf{y}_n^{\text{targ}} =[\mathbf{y}_n^{(1)}, \mathbf{y}_n^{(2)},\cdots, \mathbf{y}_n^{(R)}]\cdot \mathbf{w} \quad \forall n\in\{1,2,\cdots,N\}
\end{equation}
where $\mathbf{w}$ is set as a constant vector with all elements equal to $1/R$ in soft majority voting~\cite{bayes2_mjv}, and with variable elements that are automatically learned in WDN~\cite{wdn-aaai}. Note that in both approaches, one single $\mathbf{w}$ is shared among all $N$ samples. However, we argue that this modeling assumption can be too strong in some cases, such as the example shown in the blue box of Figure~\ref{fig:motivation}. Specifically, consider two annotators on the MNIST dataset. Because of personal writing and perception habits, the first and second annotators, AnT-1 and AnT-2, might erroneously assign label `6' to an image of `5', and label `2' to an image of  `1', respectively. Thus, for an input image of `6', the weight vector should be more biased towards the second annotator AnT-2 (say $\mathbf{w}=[0.3,0.7]^T$) since in this case the first annotator AnT-1 is more likely to provide a problematic label. Alternatively, when the input image shows `1', the weight vector should rely more on the first annotator (say with weights $\mathbf{w}=[0.7,0.3]^T$). This example implies that the weight vector should be different for different input images (i.e.,~sample-wise) when considering the reliability of annotators.

Other works such as MBEM~\cite{mbem} and TraceReg~\cite{cvpr}, instead of resorting to a weight vector $\mathbf{w}$, introduce a set of annotator confusion matrices $\{\mathbf{P}^{(r)}\in\mathbb{R}^{K\times K}\}_{r=1}^R$ to mimic the labeling processes of annotators. Specifically, the entry on the $i$-th row and $j$-th column $P_{ij}^{(r)}$ represents the probability that the $r$-th annotator returns label $i$, given that the golden label $y_n^{\star}$ equals $j$:
\begin{equation}\label{eq:confusion_matrix}
    {P}_{ij}^{(r)}=\text{Pr}[y_n^{(r)}=i\,|\,y_n^{\star}=j] \quad \forall n\in\{1,2,\cdots,N\}
\end{equation}
These works assume that the probability of the $r$-th annotator corrupting the label is independent of the input data point, and thus all samples share the same confusion matrix for the given annotator. However, similar to the case of $\mathbf{w}$, this assumption can also be too restrictive, as demonstrated in the yellow box of Fig.~\ref{fig:motivation}. This example indicates that the confusion matrix $\mathbf{P}^{(2)}$ of a single annotator AnT-2 should be different for different input images $\mathbf{x}_1$ and $\mathbf{x}_2$. Motivated by the observations that the annotator weight vector and annotator confusion matrix should be sample-wise, we propose our method in the next section.

\section{Proposed Approach}\label{sec:method}

To begin with, we assume that for each input $\mathbf{x}_n$, knowing the confusion matrix $\mathbf{P}_n^{(r)}\in \mathbb{R}^{K\times K}$ can eliminate the bias (e.g., writing or recognition habit in our example) of the $r$-th annotator and yield a clean soft label:
\begin{equation}\label{eq:y_clean}
    \mathbf{y}_n^{(r),\text{cln}}=\mathbf{P}_n^{(r)}\,\mathbf{y}_n^{(r)}
\end{equation}
where $\{\mathbf{P}_n^{(r)}\}_{r= 1}^R$ will be learned by the network. Note that in our definition of confusion matrix, the entry on the $i$-th row and $j$-th column of $\mathbf{P}_n^{(r)}$ represents the probability of $y_n^{(r),\text{cln}}=i$ given $y_n^{(r)}=j$, i.e., $\text{Pr}[y_n^{(r),\text{cln}}=i\,|\,y_n^{(r)}=j]$. Our definition differs from that of MBEM~\cite{mbem} or TraceReg~\cite{cvpr} given by Eq (\ref{eq:confusion_matrix}). We emphasize that $\mathbf{y}_n^{(r)}\in\mathbb{R}^K$ is one-hot-encoded and only one of its entries is $1$, while $\mathbf{y}_n^{(r),\text{cln}}\in\mathbb{R}^K$ is a soft stochastic vector, which can be regarded as the clean label after removing the annotator's bias. 

Once the confusion matrix $\mathbf{P}_n^{(r)}$ is learned, we can use the clean label $\mathbf{y}_n^{(r),\text{cln}}$ to guide the training of a neural network by KL divergence. In this situation, a natural thought would be to use their weighted summation with $\mathbf{w}_n\in \mathbb{R}^{R}$ as the coefficient to approximate the true label:
\begin{equation}\label{eq:y_target}
    \mathbf{y}_n^{\star}\approx\mathbf{y}_n^{\text{targ}}=[\mathbf{y}_n^{(1),\text{cln}}, \mathbf{y}_n^{(2),\text{cln}},\cdots, \mathbf{y}_n^{(R),\text{cln}}]\cdot \mathbf{w}_n
\end{equation}
where $\mathbf{w}_n$ is also learned by the network. An advantage of learning from multiple clean labels in this approach is that $\mathbf{y}_n^{\text{targ}}$ becomes more stable. Intuitively, when $\mathbf{P}_n^{(r)}$ is sufficiently good, $\{\mathbf{y}_n^{(r),\text{cln}}\}_{r=1}^R$ can be regarded as $R$ i.i.d. samples drawn from $P(Y|X=\mathbf{x}_n)$. Thus, the weighted summation of $\mathbf{y}_n^{(r),\text{cln}}$ is closer to the expected $\mathbf{y}_n^{\star}$ compared to using only one clean label, i.e., $\mathbb{E}[(\mathbf{y}_n^{\text{targ}}-\mathbf{y}_n^{\star})^2]=\frac{1}{N} \mathbb{E}[(\mathbf{
y}_n^{(1),\text{cln}}-\mathbf{y}_n^{\star})^2]$. In a nutshell, for each input $\mathbf{x}_n$, if $\mathbf{P}_n^{(r)}$ and $\mathbf{w}_n$ are available, then we can obtain $\mathbf{y}_n^{\text{targ}}$ by Eq~(\ref{eq:y_clean})-(\ref{eq:y_target}) and minimize the KL divergence between it and the neural network's label prediction $\mathbf{f}(\mathbf{x}_n)$.

\begin{figure}[!ht]
    \centering
    \includegraphics[width=1.0\textwidth]{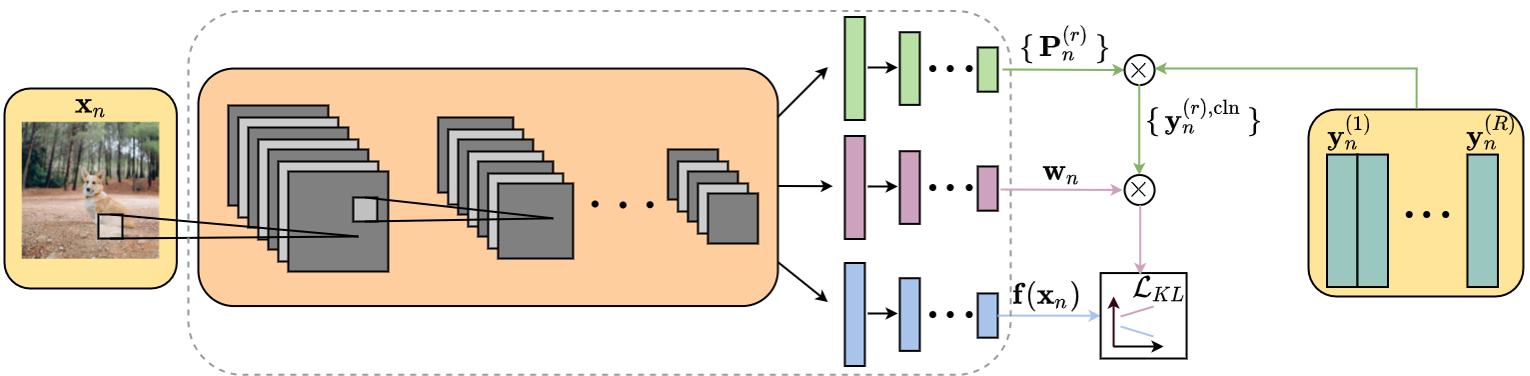}
    \caption{Training flow of our proposed method. The trainable neural network is highlighted by the dashed line. The set of confusion matrices, weight vector, and the class prediction vector share the same representation extraction network shown in the orange box. The green and purple arrows correspond to Eq (\ref{eq:y_clean}) and (\ref{eq:y_target}), respectively. During training, all paths are active, while in inference time, only the bottom blue path is active and the label prediction $\mathbf{f}(\mathbf{x}_n)$ is used.}
    \label{fig:algorithm_flow}
\end{figure}

Inspired by this idea, our model architecture and learning framework are shown in Figure~\ref{fig:algorithm_flow}. For each input $\mathbf{x}_n$, it is first fed into a deep neural network to obtain a good high-level representation, followed by three separate MLPs, outputting a set of confusion matrices $\{\mathbf{P}_n^{(r)}\}_{r= 1}^R$, a weight vector $\mathbf{w}_n$, and a class prediction $\mathbf{f}{(\mathbf{x}_n)}$, respectively. The loss function is defined as follows:
\begin{equation}
    \mathcal{L}=\frac{1}{N}\sum_{n=1}^N\mathcal{L}_{KL}(\mathbf{f}(\mathbf{x}_n),\mathbf{y}_n^{\text{targ}})
\end{equation}It should be noticed that $\mathbf{y}_n^{\text{targ}}$ is a function of $\mathbf{w}_n$ and $\{\mathbf{P}_n^{(r)}\}_{r= 1}^R$ as shown in Eq (\ref{eq:y_clean}) and (\ref{eq:y_target}). Thus, the loss function $\mathcal{L}$ is a composite function with variables $\mathbf{f}(\mathbf{x}_n)$, $\{\mathbf{P}_n^{(r)}\}_{r= 1}^R$, and $\mathbf{w}_n$, for $n=1,2,\dots,N$. These variables are what we will learn simultaneously via training the model in Figure~\ref{fig:algorithm_flow}. 

Although intuitive, further thought reveals that the above loss is insufficient. Let us provide an example of achieving minimum loss $\mathcal{L}=0$, but it is completely meaningless. Consider all entries on the first row of $\mathbf{P}_n^{(r)}$ equal to $1$ for any $n$ and $r$. Then, no matter what $\mathbf{w}_n$ is, the network always returns the prediction $\mathbf{f}(\mathbf{x}_n)=[1,0,\cdots,0]^T$ no matter what the input is. This network can achieve zero loss, but obviously it is an undesired trivial solution. The essence of the problem is that we attempt to learn the annotators' biases (and thus the clean labels) along with learning the label prediction, but the critical parameters $\{\mathbf{P}_n^{(r)}\}_{r= 1}^R$ are free to vary, which can lead to bizarre clean labels.

To overcome this, we invoke some facts about the confusion matrices so that they can be formulated as constraints and imposed on $\{\mathbf{P}_n^{(r)}\}_{r= 1}^R$. Recall in the above illustrating example, that setting the first row of all $\mathbf{P}_n^{(r)}$ equal to $1$ is to say that the clean labels always indicate the first class no matter what the annotator's labels are for all annotators on all data samples. In reality, the annotators will typically not be that bad at labeling. To encode a preference toward the provided annotator labels, we revise the loss as follows:


\begin{equation}\label{eq:loss_func}
    \mathcal{L}=\frac{1}{N}\sum_{n=1}^N\left\{\mathcal{L}_{KL}(\mathbf{f}(\mathbf{x}_n),\mathbf{y}_n^{\text{targ}})\,+\,\frac{\lambda}{R}\mathbf{u}^T\Tilde{\mathbf{D}}^2\mathbf{u}\right\}
\end{equation}
where $\Tilde{\mathbf{D}}=\text{Diag}[\mathbf{I}-\mathbf{P}_n^{(r)}]\in\mathbb{R}^{K\times K}$, $\mathbf{u}\in\mathbb{R}^K$ represents a column vector with all elements equal to 1, and $\lambda$ is a user-defined hyper-parameter. Intuitively, this quadratic term encourages the diagonal elements in $\mathbf{P}_n^{(r)}$ to be large (i.e., approaching to $1$ from below). Namely, we have implicitly assumed that annotators won't provide completely erroneous labels, considering our definition of confusion matrix. As will be visualized in the TwoMoon example, the introduction of this quadratic term prevents the occurrence of pathological $\mathbf{P}_n^{(r)}$, and thus produces a meaningful trained network. 

The training flow presented so far still has a memory problem. Namely, as shown in Figure~\ref{fig:algorithm_flow}, there are $RK^2+R+K\approx\mathcal{O}(RK^2)$ outputs for one input during training, which is orders larger than that (i.e., $\mathcal{O}(K)$) of a conventional neural network in the $K$-class classification problem. Carefully examining the training flow reveals that the memory bottleneck lies in the set of confusion matrices, which might also be a key reason why previous works have not adopted sample-wise confusion matrices. To properly address this issue, we first notice that $\mathbf{P}_n^{(r)}$ satisfies two conditions: (i)~all entries are between $0$ and $1$, and (ii)~the summation of entries in each column equals $1$. The matrix satisfying the above two requirements is known as singly stochastic~\cite{singly_double_stochastic1}. Built upon this, if the row summation also equals $1$, then the matrix is doubly stochastic~\cite{singly_double_stochastic2}. Furthermore, the Birkhoff–von Neumann theorem~\cite{birkhoff} states that any $K\times K$ doubly stochastic matrix can be decomposed into a convex combination of permutation matrices $\{\mathbf{B}_m\in\mathbb{R}^{K\times K}\}_{m=1}^M$, where $M$ is the number of basis matrices. Consequently, we impose an additional constraint that our $\mathbf{P}_n^{(r)}$ be doubly stochastic so that the theorem can be applied. Under this constraint, the network does not need to output $\mathbf{P}_n^{(r)}$, but only the coefficients $\mathbf{c}_n^{(r)}=[{c}_{n,1}^{(r)},{c}_{n,2}^{(r)},\cdots,{c}_{n,M}^{(r)}]^T\in\mathbb{R}^M$ instead. When it is needed, we can recover $\mathbf{P}_n^{(r)}$ by:
\begin{equation}\label{eq:birkhof}
    \mathbf{P}_n^{(r)} \approx {c}_{n,1}^{(r)}\mathbf{B}_1 + {c}_{n,1}^{(r)}\mathbf{B}_2 + \cdots {c}_{n,M}^{(r)}\mathbf{B}_M 
\end{equation}

\begin{figure}[!ht]
    \centering
    \includegraphics[width=1.0\textwidth]{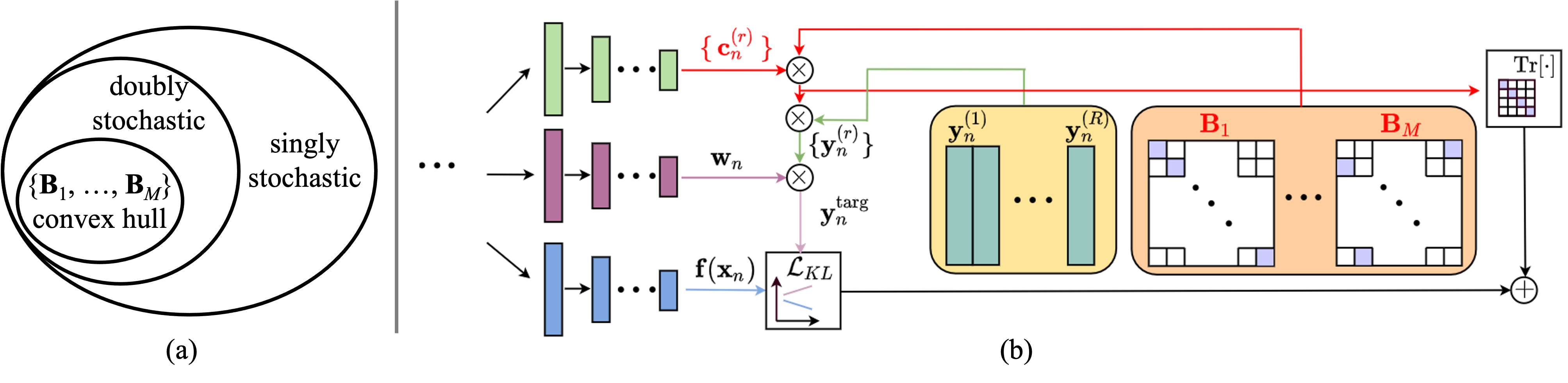}
    \caption{(a) A Venn Diagram shows the relationship among the convex hull of $\mathbf{B}_m$, doubly stochastic matrix, and singly stochastic matrix. (b) Our improved training flow. The red, green and purple arrows correspond to Eq (\ref{eq:birkhof}), (\ref{eq:y_clean}) and (\ref{eq:y_target}), respectively.}
    \label{fig:improved_flow}
\end{figure}

We employ this idea in our training flow to reduce the memory requirement, as shown in Figure~\ref{fig:improved_flow}(b). Specifically, the number $M$ of basis matrices is treated as a hyper-parameter, and the permutation matrices $\{\mathbf{B}_m\}_{m=1}^{M}$ are randomly generated before training and fixed for later use. The coefficient $\mathbf{c}_n^{(r)}$ is produced after a Softmax activation to guarantee its entries no smaller than zero and all sum to one. Essentially, we shrink the exploration space of the confusion matrix $\mathbf{P}_n^{(r)}$ as shown in Figure~\ref{fig:improved_flow}(a). Representing $\mathbf{P}_n^{(r)}$ by a convex combination of fixed permutation matrices and introducing the quadratic term into the loss function can be regarded as two techniques to regularize the confusion matrices.

\begin{figure}[!ht]
    \centering
    \includegraphics[width=0.92\textwidth]{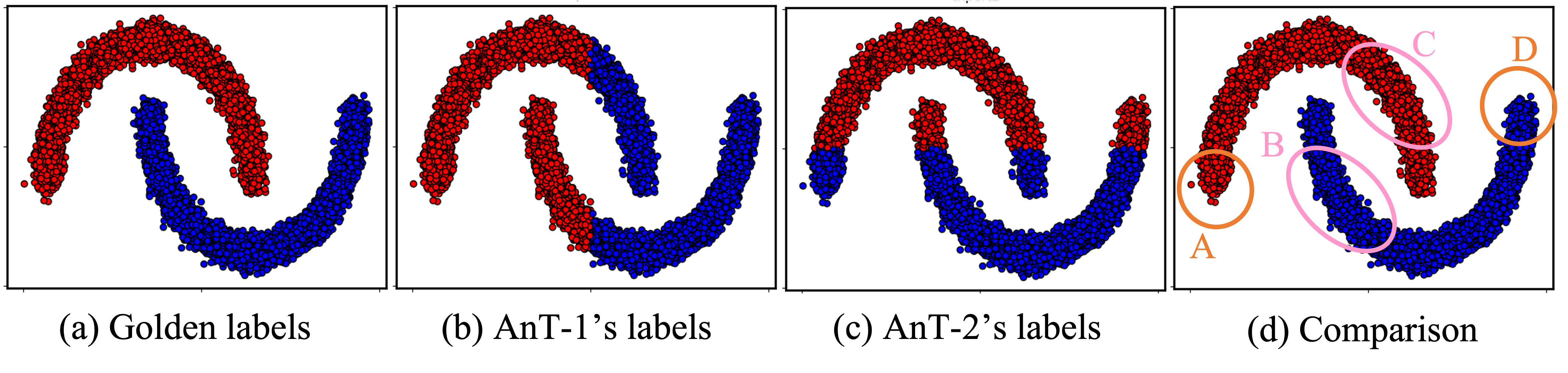}
    \caption{The modified TwoMoon toy example. Red and blue dots represent label 1 and 0, respectively. In (d), the regions where AnT-1 is wrong while AnT-2 is correct are marked with pink circles B and C. Similarly,  orange circles A and D denote where AnT-1 is correct while AnT-2 is wrong. The remaining regions are those where both annotators are wrong or correct simultaneously.}
    \label{fig:twomoonsetup}
\end{figure}

\begin{figure}[!ht]
    \centering
    \includegraphics[width=0.98\textwidth]{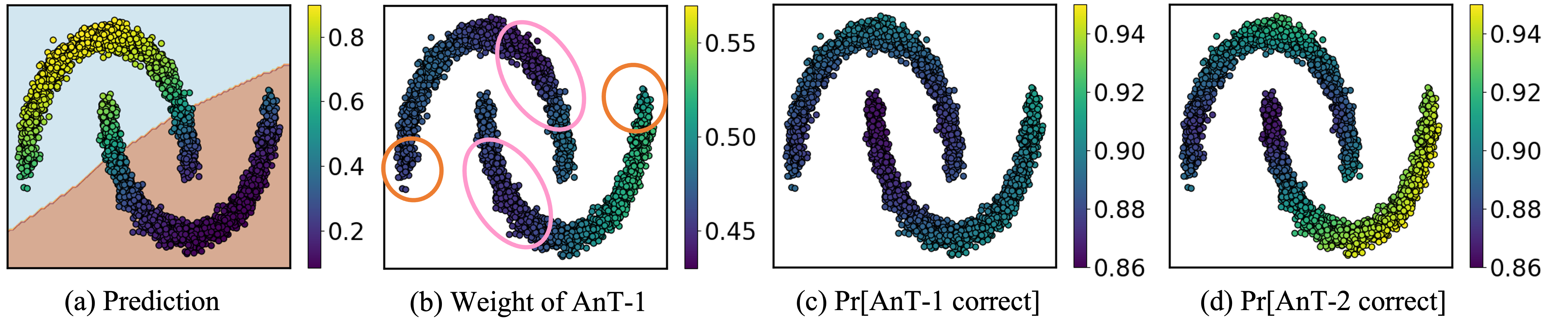}
    \caption{(a)~Prediction heatmap and classification boundary of our learned classifier. (b)~Heatmap of the first element in $\mathbf{w}_n\in\mathbb{R}^2$. (c)~Heatmap of the diagonal elements in $\mathbf{P}_n^{(1)}\in\mathbb{R}^{2\times 2}$. (d)~Heatmap of the diagonal elements in $\mathbf{P}_n^{(2)}\in\mathbb{R}^{2\times 2}$. Specifically, when the annotator's label is 0, we plot the first diagonal element, while if it is 1, we plot the second. Namely, the heatmap represents the probability that the annotator's label is correct.}
    \label{fig:twomoonheatmap}
\end{figure}

For an intuitive understanding of our approach, we perform some visualizations on the TwoMoon toy example~\cite{Xue_2021_ICCV,gao2022training}. We generate 20,000 samples on the XY-plane and specify their golden labels according to which branch (i.e., upper or lower) the samples lie in. Assume that two annotators respectively view the data from horizontal and vertical perspectives. For instance, the first annotator AnT-1 assigns label 1 to samples with X-coordinates less than zero, and label 0 to those larger than zero. See Figure~\ref{fig:twomoonsetup} for an illustration of the generated data and labels. Next, we divide data into training and test dataset according to the ratio $4:1$. In the training dataset, only annotators' labels are provided, while the golden labels are available in the test dataset. Note that TwoMoon is a binary classification task (i.e., $K=2$) and that only two $2\times2$ permutation matrices exist, so we can use two basis matrices $\{\mathbf{B}_1,\mathbf{B}_2\}$ to approximate the confusion matrix. A three-layer MLP is used as the backbone model. In one experiment running, the classifier obtained by learning from AnT-1's labels alone attains 66.51\% test accuracy, and that from AnT-2's labels alone is 83.67\%. Moreover, due to the nature of AnT-1's and AnT-2's labels, the decision boundary of these two classifiers are vertical and horizontal, respectively. On the other hand, the classifier obtained via our method achieves 86.70\% test accuracy and the decision boundary, as shown in Figure~\ref{fig:twomoonheatmap}(a), is curved, neither vertical nor horizontal anymore.

To verify that our method comes to effect, another necessary action is to examine whether the resulting weight vectors and confusion matrices are indeed sample-dependent. Figure~\ref{fig:twomoonheatmap}(b) plots the heatmap of the weight associated with AnT-1. Matching it with Figure~\ref{fig:twomoonsetup}(d), we see that the region B and C have smaller weights than $0.5$. That coincides with our intuition: because region B and C correspond to where AnT-1 is wrong while AnT-2 is correct, the weight vector should incline more to AnT-2. Applying the same reasoning, we expect region A and D are painted with light color in Figure~\ref{fig:twomoonheatmap}(b). We notice that D matches our expectation while A does not. Moreover, in Figure~\ref{fig:twomoonheatmap}(c), we do witness that the places corresponding to region B and C have darker color than that of D; while the place of region A should be light colored, it is dark in reality.  This discrepancy between our solution (Figure~\ref{fig:twomoonheatmap}) and the real case (Figure~\ref{fig:twomoonsetup}) might be attributed to two reasons: (i) the assumption of annotator labels being not too bad is violated in this example considering that there are places where both annotators are wrong, and generally (ii) the discrepancy is inevitable due to the nature of the inverse problem. Namely, the real case is only one possible scenario (or local optimum) among many that could be reached via minimizing our loss function. With our hyper-parameter setting, we currently reach the solution shown in Figure~\ref{fig:twomoonheatmap}. Nevertheless, the most important observation is that different colors (i.e., sample-dependent parameters) do indeed appear, which is our main proposition. Since our concern here is not to precisely recover the generating function, but rather to have a more accurate classifier, the example still demonstrates the potential of our method.

\section{Numerical Results}\label{sec:result}

\subsection{MNIST}
\textbf{Synthesis Method.} The MNIST dataset is a collection of handwritten digits with 10 classes~\cite{mnist}. We divide all images into a training dataset, a validation dataset, and a test dataset containing 55,000, 5,000, and 10,000 images, respectively. We note that no well-known public datasets suit our purpose (i.e., data with annotator identification) in the community, so that the existing literature all uses some sort of annotator synthesis method. For example, MBEM~\cite{mbem} and TraceReg~\cite{cvpr} use a `hammer-spammer' synthesis rule, assuming that each annotator is either a hammer, always returning golden labels, or a spammer, randomly choosing labels in a  uniform way. However, it appears to us that bad annotators are generally better than randomly picking labels and good annotators sometimes also make mistakes. Thus, in our experiment, we synthesize three annotators according to the following rule:  We assume that for each of these annotators, there exists one `weakness' image that the annotator cannot identify. For the images whose Euclidean distance to the `weakness' image are smaller than a threshold $\epsilon$, the annotator will randomly pick another class label different from the golden one, in a uniform way. Otherwise, the annotator will provide the correct label. 

\paragraph{\textbf{\textup{Setting.}}} For comparison purpose, besides training with solely one single annotator's labels, we also implement majority voting, TraceReg ~\cite{cvpr}, MBEM~\cite{mbem}, and WDN~\cite{wdn-aaai} as baselines. We choose LeNet~\cite{mnist} as the backbone model, set the number of epochs to $40$, learning rate to $0.01$, and use SGD with momentum as the optimizer. For our method, we randomly generate 20 (i.e.,~$M=20$) base permutation matrices to approximate the confusion matrices, and set the hyper-parameter $\lambda$ in Eq.~(\ref{eq:loss_func}) to $1.0$. Following the setting of TraceReg~\cite{cvpr}, we assume that golden labels are available in the validation dataset, so that validation can be used to compare models among different training epochs and select the optimal model for a specific method. 

\begin{table}[!ht]
\small
\centering
\caption{Accuracies (\%) on MNIST under different annotator skills $\epsilon$}

\begin{tabular}{p{2.5cm}p{1.5cm}p{1.5cm}p{1.5cm}p{1.5cm}p{1.5cm}p{1.5cm}}
\toprule
& $\epsilon= 30$ & $\epsilon = 31$ &  $\epsilon = 32$  &  $\epsilon = 33$ & $\epsilon=34$ & $\epsilon=35$\\
\midrule
w/~AnT-1's  & 80.79          & 75.67          & 67.48          & 59.49           & 47.63        & 43.06   \\
w/~AnT-2's  & 66.53          & 57.47          & 49.25          & 41.42           & 33.68        & 25.99   \\
w/~AnT-3's  & 48.35          & 40.75          & 33.01          & 27.08           & 21.25        & 16.25  \\
Mjv         & 87.72          & 80.35          & 74.47          & 67.15           & 56.75        & 44.46   \\
TraceReg    & 85.98          & 76.91          & 70.47          & 62.38           & 51.52        & 40.09  \\
MBEM        & 83.30          & 74.66          & 66.35          & 56.52           & 49.36        & 36.65  \\
WDN         & 81.09          & 72.16          & 64.32          & 52.90           & 44.50        & 32.00  \\
Ours        & \textbf{92.49} & \textbf{80.67} & \textbf{76.76} & \textbf{70.44}  & \textbf{61.19} & \textbf{46.21}\\
\midrule
w/~true label & 99.20          & 99.20           & 99.20          & 99.20            & 99.20  & 99.20\\ 
\bottomrule
\end{tabular}
\label{table:MNIST}
\end{table}

\begin{figure}[!ht]
    \centering
    \includegraphics[width=0.9\textwidth]{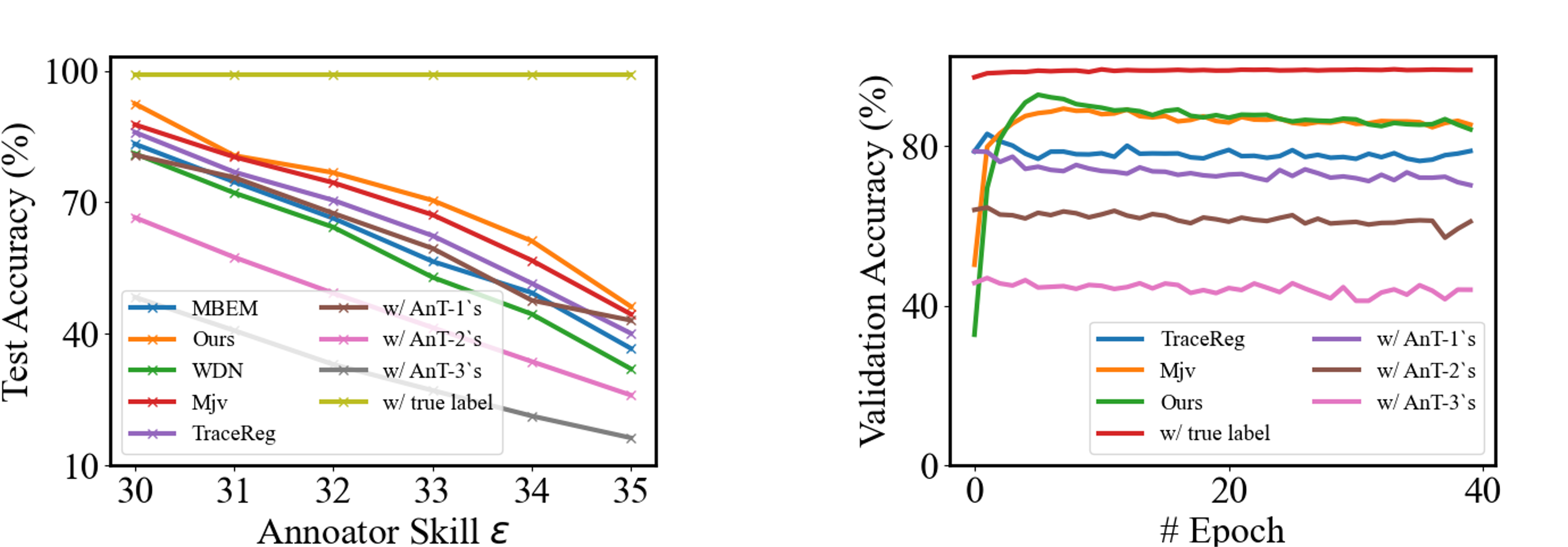}
    \caption{MNIST test cases. Left: Test accuracy is plotted versus different annotator skills. Right: During the training phase of one experiment with $\epsilon=30$, the validation accuracy is plotted as a function of the number of epochs. Since MBEM~\cite{mbem} has an additional loop (corresponding to EM algorithm) outside the training of classifier, it has been omitted from the figure. WDN~\cite{wdn-aaai} is omitted for a similar reason.}
    \label{fig:mnist_result}
\end{figure}

\noindent\textbf{Results.} The accuracies of different optimal models yielded by different methods on the test dataset are reported in Table~\ref{table:MNIST}. All results are reported by averaging five independent experiments, and the left plot of Figure~\ref{fig:mnist_result} visualizes these results. Our method outperforms all other methods and is the closest to training with true golden labels (i.e., the upper bound) under all different annotator skills. Moreover, at $\epsilon=30$, our method achieves $4.77\%$ more accuracy compared to the second best method. It is worth mentioning that in some cases, TraceReg~\cite{cvpr} and MBEM~\cite{mbem} perform even worse than majority voting, which seems contradictory to what they report. However, this is because our annotator synthesis method is different from theirs, and we hypothesize that their method might not perform well under a sample-dependent synthesis (as in our case) due to the assumption of sample-independent weight vector or confusion matrix. We perform further experiments under their synthesis method; for this and further discussion, please refer to the supplementary material. 

In the right plot of Figure~\ref{fig:mnist_result}, we show model accuracy on the validation dataset as a function of the number of epochs. As classification on MNIST is rather easy, all methods converge quickly after only a few epochs. We see that our method is almost consistently better than all other methods among all epochs.

\subsection{CIFAR-100}

\textbf{Setting.} CIFAR-100~\cite{cifar} is a collection of $32\times 32$ RGB images with 100 classes. We divide all images into a training dataset, a validation dataset and a test dataset containing 40,000, 10,000, and 10,000 images, respectively. Following the synthesize method described in the previous subsection, we create three annotators AnT-1, AnT-2 and AnT-3. We select ResNet18 as the backbone model. Note that we have reduced the convolution kernel size to $3\times 3$ to suit the case of $32\times 32$ RGB input. We choose SGD with momentum as the optimizer, set learning rate to 0.1, and number of epochs to 200. For our method, the hyper-parameter $\lambda$ is set to $1.0$, and since now the number of classes $K$ is 100, we randomly generate 150 (i.e.,~$M=150$) base permutation matrices to approximate the confusion matrix.

\begin{figure}[!ht]
    \centering
    \includegraphics[width=0.9\textwidth]{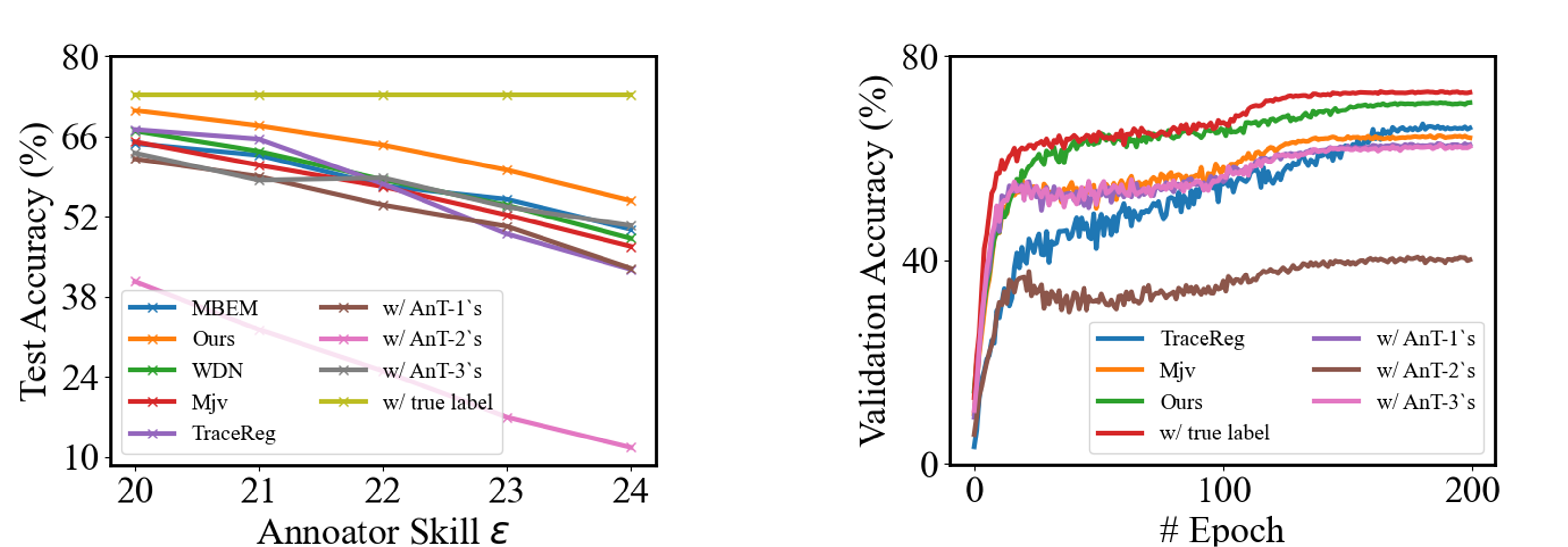}
    \caption{CIFAR-100 test cases. Left: Test accuracy is plotted versus different annotator skills. Right: During the training phase of one experiment with $\epsilon=20$, the validation accuracy is plotted as a function of the number of epochs.}
    \label{fig:cifar_result}
\end{figure}

\paragraph{\textbf{\textup{Results.}}} Table~\ref{table:exp_cifar100} reports the test accuracies of different methods on CIFAR-100 under different expert skills $\epsilon$. With $\epsilon$ increasing, the annotators' labels become worse and all methods' accuracies drop. Moreover, it is interesting to note that when $\epsilon=24$, the test accuracy obtained with a model trained solely on AnT-3's labels is better than all other baselines except our method. As shown in the left plot of Figure~\ref{fig:cifar_result}, our method always achieves the highest accuracy among all other baselines under different $\epsilon$. For instance, when $\epsilon =22$, our method attains $5.74\%$ more accuracy compared to the second best method. Moreover, as shown in the right plot of Figure~\ref{fig:cifar_result}, our method (i.e.,~green line) is consistently better than other baselines after about 30 epochs, and it is even comparable to training with true labels (i.e.,~red line) at around 50 to 100 epochs. A few ablation studies (such as varying $\lambda$, only using $\mathbf{w}_n$ or $\mathbf{P}_n^{(r)}$, using pretrained neural networks as annotators) are performed in the supplementary.

\begin{table}[!ht]
\small
\centering
\caption{Accuracies (\%) on CIFAR-100 under different expert skills $\epsilon$}
\begin{tabular}{p{3cm}p{1.5cm}p{1.5cm}p{1.5cm}p{1.5cm}p{1.5cm}}
\toprule
& $\epsilon = 20$ & $\epsilon = 21$ &  $\epsilon = 22$  &  $\epsilon = 23$ & $\epsilon=24$\\
\midrule
w/~AnT-1’s    & 62.05 & 58.98 & 54.02 & 50.27 & 42.97 \\ 
w/~AnT-2’s    & 40.64 & 32.26 & 25.04 & 16.99 & 11.67 \\ 
w/~AnT-3’s    & 63.07 & 58.33 & 58.74 & 53.68 & 50.42 \\ 

Mjv             & 65.06 & 60.97 & 57.22 & 52.25 & 46.77 \\ 
TraceReg        & 67.16 & 65.51 & 57.70 & 48.96 & 42.79 \\ 
MBEM            & 64.75 & 62.69 & 57.37 & 55.02 & 49.74 \\ 
WDN             & 66.92 & 63.37 & 58.42 & 53.97 & 48.17 \\ 
Ours            & \textbf{70.05} & \textbf{67.83} & \textbf{64.48} & \textbf{60.12} & \textbf{54.77} \\ 
\midrule
w/~true label  & 73.24 & 73.24 & 73.24 & 73.24 & 73.24 \\ 
\bottomrule
\end{tabular}
\label{table:exp_cifar100}
\end{table}

\subsection{ImageNet-100}

\textbf{Setting.} To make the total running time affordable, we conduct the experiment on a well-known subset of ImageNet, ImageNet-100, in this example. It is a subset of ImageNet dataset~\cite{lifeifei-imagenet}, containing 100 random classes and a total of 135,000 images. We divide all images into a training dataset, a validation dataset and a test dataset with ratio of 25:1:1. Following the synthesize method described in previous section, we create three annotators AnT-1, AnT-2, and AnT-3. We choose ResNet18 as the backbone model and SGD with momentum as the optimizer. We set learning rate to 0.1 and number of epoch to 200. For our method, we set the hyper-parameter $\lambda$ to $1.0$ and randomly generate $M=150$ permutation matrices as bases. 

\begin{figure}[!ht]
    \centering
    \includegraphics[width=0.9\textwidth]{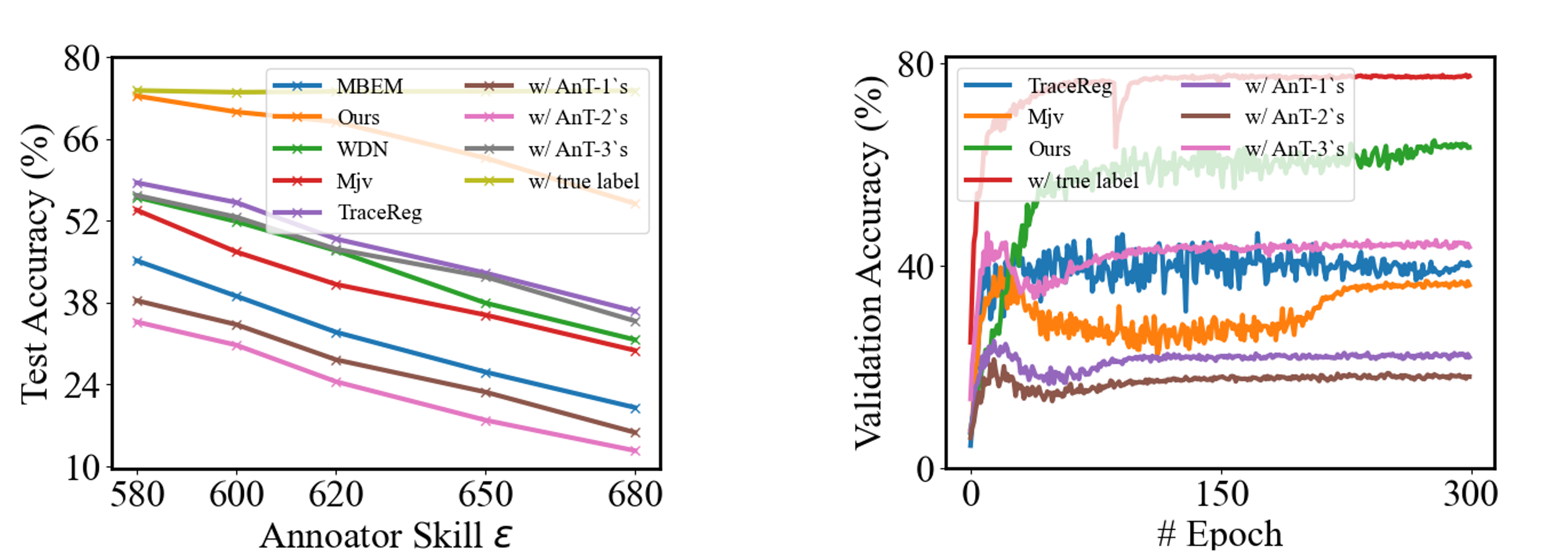}
    \caption{ImageNet-100 test cases. Left: Test accuracy is plotted versus different annotator skills. Right: During the training phase of one experiment with $\epsilon=20$, the validation accuracy is plotted as a function of the number of epochs.}
    \label{fig:imagenet_result}
\end{figure}

\paragraph{\textbf{\textup{Results.}}} Table~\ref{table:exp_imagenet100} reports the test accuracy of different methods on ImageNet-100 under expert skill $\epsilon$. Our method outperforms all baselines under this setting, and is consistently better than other baselines after around 50 epochs as shown in Fig.~\ref{fig:imagenet_result}. Moreover, in this example, we observe that our method achieves more accuracy improvement compared to that in MNIST or CIFAR-100. 

\begin{table}[!ht]
\small
\centering
\caption{Accuracies (\%) on ImageNet-100 under different expert skills $\epsilon$}
\begin{tabular}{p{3cm}p{1.5cm}p{1.5cm}p{1.5cm}p{1.5cm}p{1.5cm}}
\toprule
& $\epsilon = 580$ & $\epsilon = 600$ &  $\epsilon = 620$  &  $\epsilon = 650$ & $\epsilon=680$\\
\midrule
w/~AnT-1’s      & 38.36 & 34.25 & 28.24 & 22.71 & 15.81 \\ 
w/~AnT-2’s      & 34.72 & 30.75 & 24.50 & 17.91 & 12.70 \\ 
w/~AnT-3’s      & 56.35 & 52.62 & 47.16 & 40.41 & 34.86 \\ 

Mjv             & 53.83 & 46.67 & 41.13 & 35.90 & 29.83 \\ 
TraceReg        & 58.51 & 55.13 & 48.90 & 43.07 & 36.54 \\ 
MBEM            & 45.18 & 39.12 & 32.90 & 26.14 & 20.06 \\ 
WDN             & 56.03 & 51.86 & 46.90 & 37.94 & 31.65 \\ 
Ours            & \textbf{73.34} & \textbf{70.59} & \textbf{68.91} & \textbf{62.72} & \textbf{54.93} \\ 
\midrule
w/~true label  & 74.26 & 74.26 & 74.26 & 74.26 & 74.26 \\ 
\bottomrule
\end{tabular}
\label{table:exp_imagenet100}
\end{table}

\section{Conclusions and Future Work}\label{sec:conclusion}

\noindent\textbf{Limitations and Future Work.}
Here we discuss the limitations of our method and potential future work. First, even with the matrix decomposition method, the number of our neural network outputs is still on the order of $\mathcal{O}(RM)$. Although we can choose $M\sim\mathcal{O}(K)$, the complexity remains linear in $R$. This is not a problem when $R$ is small (e.g., $R=3$ in our experiments), while complexity may be a concern when there are many annotators. However, MBEM~\cite{mbem} has demonstrated its validity in this situation. Future work could explore extensions to our proposed method for the large $R$ case.

In addition to scalability, another direction worth exploration is dealing with missing annotator labels, as MBEM~\cite{mbem} and TraceReg~\cite{cvpr} do. Namely, an annotator might not provide labels for all data samples, and sometimes some labels might be missing. The MBEM~\cite{mbem} and TraceReg~\cite{cvpr} frameworks implicitly handle this situation. However, it is less clear how our method could efficiently address such scenarios. One possible approach would be using a mask operation in Eq.~(\ref{eq:y_clean}). Namely, we could add a vector $\mathbf{s}_{n}^{(r)}\in\mathbb{R}^K$ ahead of $\mathbf{P}_n^{(r)}$ on the right hand side of the equation. If the $r$-th annotator does not provide a label for the $n$-th data, then the elements of $\mathbf{s}_{n}^{(r)}$ are set to all zeros, otherwise, all ones. However, this method might be inefficient if each annotator only provides labels for a small subset of all images (i.e., annotators' labels are sparse). A related avenue is to explore if we could use such sparsity to address the scalability issue.

\paragraph{\textbf{\textup{Conclusions.}}} In this paper, we propose a novel method to learn a classifier given a noisy training dataset, in which each data point has several labels from multiple annotators. Our key idea is to make the weight vectors and the confusion matrices data-dependent. Moreover, we realize two regularization methods for the confusion matrix to guide the training process: one is to include a quadratic term inside the loss function, and the other is to confine the confusion matrix as a convex combination of permutation matrices. Our visualization on the TwoMoon dataset verifies that the learned parameters are indeed sample-wise, and our numerical results on MNIST, CIFAR-100 and ImageNet-100 demonstrate that our method outperforms various state-of-the-art methods.

\paragraph{\textbf{\textup{Acknowledgements.}}} This research was supported in part by Millennium Pharmaceuticals, Inc. (a subsidiary of Takeda Pharmaceuticals). The authors also acknowledge helpful feedback from the reviewers.
Zhengqi Gao would like to thank Alex Gu, Suvrit Sra, Zichang He and Hangyu Lin for useful discussions, and Zihui Xue for her support.

\bibliographystyle{splncs04}
\bibliography{egbib}

\newpage

\section*{Supplementary Material}

\section*{A~Ablation Studies on MNIST}

\textbf{Hammer-Spammer Synthesis.} Here we conduct experiments on MNIST under the hammer-spammer synthesis method used in MBEM and TraceReg. Specifically, we consider the class-wise hammer-spammer synthesis. A hammer refers to always correct and a spammer refers to always wrong. In their original implementations, an annotator is a hammer with probability $p$ and a spammer with probability $1-p$ for any class $k\in\{0,1,\cdots,K-1\}$ (where $K=10$ in MNIST dataset). During our implementation, we find that this synthesis has too much variance. Namely, one annotator provides wrong labels for all samples while other annotators provide correct labels for some classes and achieve moderate accuracies. Thus, we slightly revise the hammer-spammer implementation. For one annotator, we randomly select $N_{\text{correct}}$ classes and specify that this annotator is a hammer on those classes, and a spammer on the remaining ones. In this example, we make up five annotators following our class-wise hammer-spammer synthesis. Other settings (e.g., learning rate, hyper-parameter) are identical to the experiment in the main text.

\begin{table}[!ht]
\small
\centering
\caption{Test accuracies of different methods (\%) are evaluated under $N_{\text{correct}}=3$. The test accuracy of training with golden labels (i.e., upper bound) is 99.20\%.}

\begin{tabular}{p{1.8cm}<{\centering}p{1.8cm}<{\centering}p{1.6cm}<{\centering}p{1.6cm}<{\centering}p{1.1cm}<{\centering}p{1.2cm}<{\centering}p{1.cm}<{\centering}p{1.cm}<{\centering}}
\toprule
Max AnTs' &  Mean AnTs' &  Min AnTs' & TraceReg & Mjv  & MBEM & WDN & Ours\\
\midrule
40.68\% & 39.56\% & 38.55\% & 84.54\% & 86.27\%   & \textbf{88.31\%} & 88.26\% & {88.26\%}\\
\bottomrule
\end{tabular}
\label{table:exp_mnist_supp}
\end{table}

Test accuracies of different methods are reported in Table \ref{table:exp_mnist_supp}. The first, second, and third columns report the max, mean, and min test accuracy that the model trained solely with one annotator's labels could achieve. All values in the table are reported after averaging five independent experiments. We observe that in this case, MBEM, WDN, and our method achieve similar accuracies and outperform others. Since in this setting, the annotator synthesis method is not sample-wise\footnote{Precisely, not as sample-wise as the synthesis method in our main text.}, it makes sense that our method is not outstanding. This alternatively indicates that our method might be better suitable to the case when the annotator labels are data-dependent. When labels are data-independent, methods such as MBEM might be sufficient. Nevertheless, it appears to us that in real applications, sample-wise annotator labels seem more reasonable. It would be great if a public dataset is available in the future so that all methods could be tested on the same page.

\paragraph{\textbf{\textup{Remarks on our Euclidean synthesis.}}} \emph{At present, no publicly available dataset provides annotator error data, and we have to synthesize annotators' labels in some way}. In previous literature, they also use synthetic methods (e.g., the hammer-spammer synthesis mentioned above) to generate annotator labels on ImageNet, CIFAR-10. We believe a synthesis method based on image similarity metric is more realistic, and Euclidean distance is widely used in literature and code packages. Another major reason to use Euclidean distance in synthesis is that it can measure similarity not only in images, but also in audio, text, etc. Our Euclidean distance is calculated on raw inputs. Using it on latent features would require training a good network and performing inference on all data in preprocessing, which is time consuming. A dataset containing multiple annotators' labels for one data sample would greatly help the community, as we could then more easily inspect the performances of different methods with apple-to-apple comparisons. 

\paragraph{\textbf{\textup{Remarks on the range of $\epsilon$ in our synthesis.}}} Going back to our annotator label synthesis methods, when Euclidian distance between an image and annotator's `weakness' image is smaller than $\epsilon$, random labels are returned. The range of $\epsilon$ (e.g., $[30,35]$ in Table~1 of the main text) was selected as follows: (i) min $\epsilon$ corresponds to when annotators provide wrong labels, but our method achieves similar accuracy as training with true labels; (ii) max $\epsilon$ corresponds to training where one annotator's labels gives disastrous accuracy (e.g., around $15\%$).
Case~(i) shows how bad the annotators' labels can be such that our proposed method will start to drop from the golden accuracy. Case~(ii) demonstrates how good our method is when at least one annotators' labels are almost completely unreliable. Our choices of min and max $\epsilon$ lie at two extreme ends and the chosen range is wide enough to verify the method at all cases. Similar criterion is applied to CIFAR-100 and ImageNet-100 experiments.

\section*{B~Ablation Studies on CIFAR-100}

\begin{figure}[!htb]
    \centering
    \includegraphics[width=0.9\linewidth]{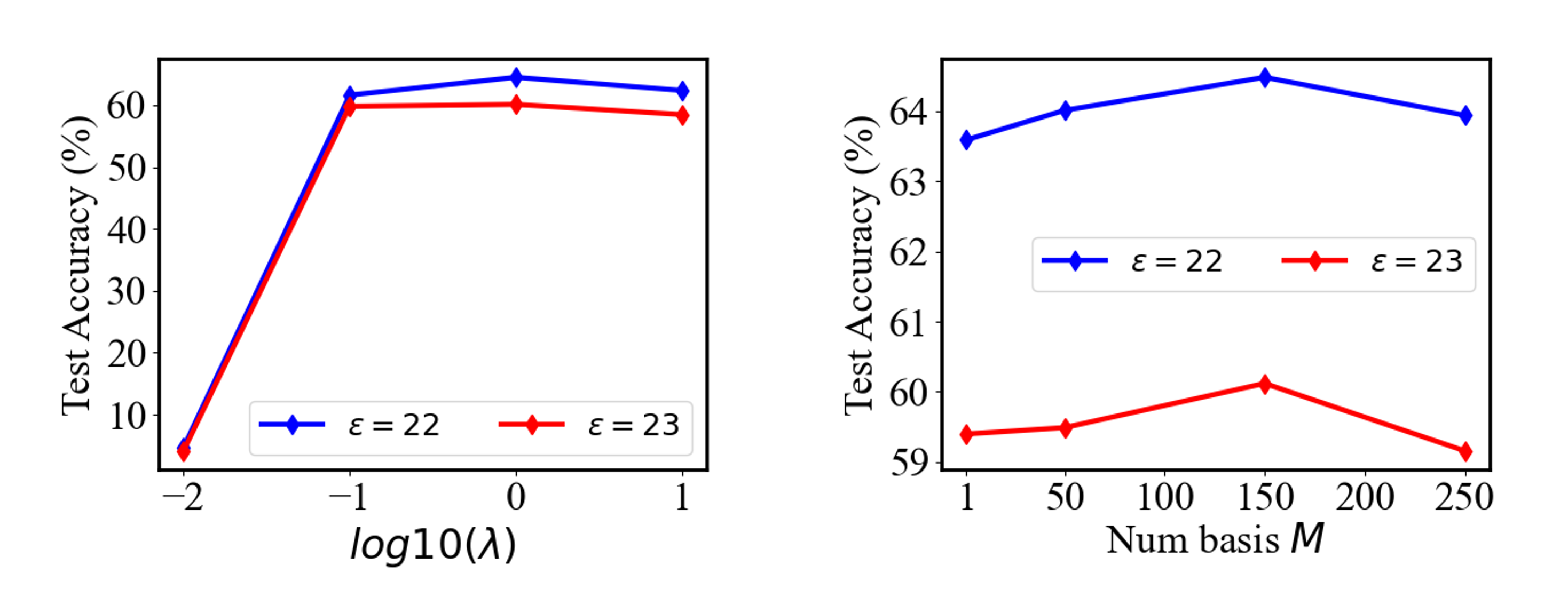}
    \caption{Left: ablation study on $\lambda$. The best test accuracy is achieved at $\lambda=1.0$ for both $\epsilon=22$ and $\epsilon=23$. This is also the value we use in the main text. Right: ablation study on $M$. The best test accuracy is achieved at $M=150$.}
    \label{fig:ablation_hyperlambda_basism}
\end{figure}

Here we report results of ablation studies on (i)~only using confusion matrices $\{\mathbf{P}_n^{(r)}\}_{r= 1}^R$, (ii)~only using weight vectors $\mathbf{w}_n$, (iii)~hyper-parameter $\lambda$, and (iv)~number of basis matrices $M$ in Table \ref{table:exp_cifar100_supp} and Fig.~\ref{fig:ablation_hyperlambda_basism}.

\begin{table}[!ht]
\centering
\caption{Accuracies (\%) on CIFAR-100 under different $\epsilon$.}
\begin{tabular}{p{1.2cm}<{\centering}p{3.0cm}<{\centering}p{3.0cm}<{\centering}p{3.0cm}<{\centering}}
\toprule
$\epsilon$ & Ours (w/ only $\mathbf{w}_n$) & Ours (w/ only $\mathbf{P}_n^{(r)}$) & Ours (w/ both) \\
\midrule
$22$  & 63.59 & 60.25 & \textbf{64.48} \\
$23$  & 59.40 & 57.07 & \textbf{60.12} \\
\bottomrule
\end{tabular}
\label{table:exp_cifar100_supp}
\end{table}

As we mentioned earlier, there is  no publicly available dataset 
provides annotator error data, and we have to synthesize annotators' labels in some way. Most of our experiments are carried out based on our Euclidean synthesis technique. Here we perform an extra experiment using neural networks as annotators in Table~\ref{table:exp_cifar100_supp2}. This, to the best of our efforts, is the closest to a real-world scenario. Specifically, we inspect the reported test accuracies of the trained models listed at \href{https://github.com/chenyaofo/pytorch-cifar-models}{https://github.com/chenyaofo/pytorch-cifar-models}. Next, for our experimental purpose, we deliberately choose three models (MobileNet, Vgg11, ShuffleNet) which perform badly, download and regard them as the three annotators. Results are reported in the following Table 3. 

\begin{table}[!ht]
\small
\centering
\caption{Accuracies (\%) on CIFAR-100. We take pretrained MobileNet, Vgg11, ShuffleNet as the three annotators. Their test accuracies are 65.28\%, 66.90\%, 60.17\%, respectively. We note that since our data normalization might be different from the one used to originally train these three models, we witness a difference between the accuracies reported here and on \href{https://github.com/chenyaofo/pytorch-cifar-models}{https://github.com/chenyaofo/pytorch-cifar-models}. The results of this experiment show that our label fusion approach again outperforms other annotator error methods, with this alternative setup to obtain example annotator errors.}
\begin{tabular}{p{1.8cm}<{\centering}p{1.8cm}<{\centering}p{1.8cm}<{\centering}p{1.8cm}<{\centering}p{1.8cm}<{\centering}|p{1.8cm}<{\centering}}
\toprule
MBEM &  WDN & TraceReg & Mjv & Ours & w/ true\\
\midrule
70.48 & 72.08 & 71.47 & 70.60 & \textbf{73.14} & 73.24 \\
\bottomrule
\end{tabular}
\label{table:exp_cifar100_supp2}
\end{table}

\end{document}